\documentclass[10pt,twocolumn,letterpaper]{article}

\usepackage{cvpr}              
\usepackage{cuted}   
\usepackage{capt-of} 
\definecolor{cvprblue}{rgb}{0.21,0.49,0.74}
\usepackage[pagebackref,breaklinks,colorlinks,allcolors=cvprblue]{hyperref}
\usepackage[table]{xcolor}
\usepackage{comment}

\def\paperID{16817} 
\def\confName{CVPR}
\def\confYear{2026}

\title{Alpha Divergence Losses for Biometric Verification}

\author{Dimitrios Koutsianos$^{1,2}$ \hspace{0.5em} Ladislav Mosner$^3$\hspace{0.5em} Yannis Panagakis$^{2,4}$ \hspace{0.5em} Themos Stafylakis$^{1,2,5}$
\\
$^1$Athens University of Economics and Business, Greece \hspace{0.5em} $^2$Archimedes/Athena RC, Greece \hspace{0.5em} \\
$^3$Brno University of Technology, Czechia\\
$^4$National and Kapodistrian University of Athens, Greece \\
$^5$Omilia - Conversational Intelligence, Greece\\
{\tt\small \{dkoutsianos,tstafylakis\}@aueb.gr}}

\newcommand{\methodB}{\textsc{A3M}} 

\begin{document}
\maketitle
\begin{abstract}
Performance in face and speaker verification is largely driven by margin-based softmax losses such as CosFace and ArcFace. Recently introduced $\alpha$-divergence loss functions offer a compelling alternative, particularly due to their ability to induce sparse solutions (when $\alpha>1$). However, integrating an angular margin—crucial for verification tasks—is not straightforward. We find that this integration can be achieved in at least two
distinct ways: via the reference measure (prior probabilities) or via the logits (unnormalized log-likelihoods). In this paper, we explore both pathways, deriving two novel margin-based $\alpha$-divergence losses: Q-Margin (margin in the reference measure) and A3M (margin in the logits). We identify and address a training instability in A3M—caused by sparsity—with a simple yet effective prototype re-initialization strategy.
Our methods achieve significant performance gains on the challenging IJB-B and IJB-C face verification benchmarks. We demonstrate similarly strong performance in speaker verification on VoxCeleb. Crucially, our models significantly outperform strong baselines at low false acceptance rates (FAR). This capability is critical for practical high-security applications, such as banking authentication, where minimizing false authentications is paramount. Finally, the sparsity of the $\alpha$-divergence-based posteriors enables memory-efficient training, which is crucial for datasets with millions of identities.
\end{abstract}    
\section{Introduction}
\label{sec:intro}

Margin-based softmax losses such as CosFace~\cite{cosface} and 
ArcFace~\cite{arcface} have become the de facto standard in modern 
face and speaker recognition. They achieve state-of-the-art accuracy by incorporating 
margin penalties to the logit of the ground truth identity, which explicitly maximizes inter-class 
separation. Such methods rely on the $\operatorname{softargmax}$ operator, which 
transforms logits into probabilities evaluated by the cross-entropy 
loss or equivalently, the Kullback–Leibler (KL) divergence with a uniform 
prior~\cite{alpha-loss}. To date, margin penalties have been applied 
exclusively through geometric modifications to logits (\eg, 
$\cos(\theta + m)$ in ArcFace, $\cos(\theta) - m$ in CosFace). 
While effective, these geometric margins are heuristic extensions of 
softmax that remain decoupled from its probabilistic formulation.

Recent work by Roulet et al.~\cite{alpha-loss} introduced a principled 
generalization of the standard cross-entropy loss through Fenchel–Young 
losses~\cite{fenchel-young-losses} derived from 
$f$-divergences~\cite{f-divergence}. This framework extends cross-entropy 
in two key ways: (1)~by replacing the KL divergence with a broader family 
of $f$-divergences, each defining a unique convex loss and corresponding 
probability mapping (the $\operatorname{softargmax}_f$), and (2)~by allowing non-uniform 
reference measures $\mathbf{q}$ to encode arbitrary prior probabilities. 
While Roulet et al.\ used non-uniform $\mathbf{q}$ to represent class 
frequency imbalance, no prior work has exploited this flexibility to 
encode margin penalties probabilistically.

Among the $f$-divergence family, the $\alpha$-divergence is 
particularly useful for biometric verification due to its tunable 
trade-off between smooth and sparse probability mappings. It defines a 
continuum of generalized entropy measures: recovering the Shannon 
negentropy (standard cross-entropy) as $\alpha \rightarrow 1$, the Gini 
negentropy (sparsemax~\cite{sparsemax}) at $\alpha = 2$, and exhibiting 
intermediate behaviors for values such as $\alpha = 1.5$. For $\alpha > 1$, the $\alpha$-$\operatorname{softargmax}$ suppresses small probabilities,
producing sparse posteriors that emphasize confident predictions, which
in practice encourages more discriminative embeddings and sharper decision
boundaries in face and speaker recognition.

However, combining $\alpha$-divergence with margin-losses can be challenging: the loss's sparsity-inducing gradients can  
effectively eliminate the learning signal for certain identities, such as those with few training examples. This 
occurs because the $[\cdot]_+$ operator in the gradient formulation (defined later in the paper)
zeroes out probabilities when logits fall below a threshold. This can result in misalignment between certain prototypes and embeddings, resulting in $p_y=0$, where $y$ the ground-truth identity. Importantly, this misalignment arises primarily during the 
early stages of training, when identity prototypes are poorly aligned and 
target logits are small. Once prototypes are well-formed, the gradients 
remain stable even under geometric margins—a property exploited by our 
second method.

We address this challenge through two approaches that 
successfully unify $\alpha$-divergence with margin-based learning. Our 
first method, Q-Margin, introduces a \emph{probabilistic margin} by 
encoding the penalty directly in the reference measure $\mathbf{q}$. 
By down-weighting the ground-truth class while keeping other classes 
uniform, it avoids the instability at the loss level. 

Our second method, \methodB-I, retains the conventional \emph{geometric 
margin} formulation but stabilizes training through a mid-training 
prototype re-initialization. After several epochs, we recompute 
normalized identity prototypes from feature embeddings and replace the 
classifier weights, realigning the decision boundaries when sparse 
gradients would otherwise cause misalignment. Although not probabilistic, 
\methodB-I complements Q-Margin by addressing optimization-level 
instability.

By reinterpreting margin learning through the lens of $\alpha$-divergence, 
we unify geometric and probabilistic perspectives within a single 
framework that is both theoretically grounded and practically stable, 
bridging a fundamental gap between information-theoretic loss design 
and the empirical success of margin-based softmax. Together, these two 
strategies make $\alpha$-divergence losses practical for large-scale 
face and speaker recognition without sacrificing their theoretical advantages.

\textbf{Our contributions are as follows:}
\begin{itemize}
    \item We propose $\alpha$-divergence losses with margin penalties 
    as an inherently-sparse alternative to standard logistic loss, for training highly discriminative identity embeddings.
    
    \item We propose Q-Margin, which introduces a probabilistic margin 
    through non-uniform reference measures ($q_y = \exp(-s \cdot m)$), 
    which recovers CosFace 
    as $\alpha \rightarrow 1$.
    
    \item We propose \methodB-I, which stabilizes training through 
    mid-training prototype re-initialization, addressing optimization-level 
    misalignment for large and noisy datasets such as WebFace42M.

    \item We achieve consistent improvements over strong baselines on 
    IJB-B and IJB-C, with Q-Margin showing larger gains at 
    FRR@FAR=$10^{-4}$ and \methodB-I excelling at the stricter 
    $10^{-5}$ threshold.
\end{itemize}

\section{Related Work}
\label{sec:related_work}

\subsection{Advances in Face Recognition: Margin Losses and Classifier Efficiency}

Early deep face recognition models utilized the standard Softmax loss \cite{VGGFace2}, but this approach was found to be insufficient for learning features with enough discriminative power for open-set recognition tasks. This limitation spurred the development of loss functions that explicitly engineer the feature space to increase inter-class variance while minimizing intra-class variance.

The introduction of a margin into the loss function was a pivotal moment. L-Softmax \cite{L-Softmax}, introduced by Liu et al., was a pioneering effort that introduced an angular margin, forcing decision boundaries to be more compact. This concept was further refined by SphereFace \cite{sphereface}, which normalized the final layer's weights and constrained the learned features to lie on a hypersphere, applying a multiplicative angular margin. However, these methods proved challenging to train. 

The field of Face Recognition subsequently shifted towards additive margins, which proved to be more stable and effective. CosFace \cite{cosface} proposed adding a margin directly in the cosine space ($cos(\theta) - m$), while ArcFace~\cite{arcface} added an angular margin directly to the angle itself ($cos(\theta + m)$). ArcFace in particular has become a powerful and widely used baseline. Inspired by the great success of margin-based loss functions in face recognition, CosFace and ArcFace have been extensively adopted in the field of speaker verification~\cite{xiang2019_margin-matters}. These approaches have proven to be crucial components for enhancing system performance, as demonstrated by their consistent use in multiple VoxSRC challenges~\cite{huh2024_voxsrc-retrospective}. The concept of margin paired with a controllable difficulty of examples by duration evolved into widely used large margin fine-tuning~\cite{Thienpondt21:lm-finetuning}, allowing for the refinement of pre-trained speaker embedding extractors using much higher margins.

While effective, the fixed-margin approach assumes that all identities and samples can be learned with the same degree of difficulty. Recent research has challenged this, leading to a new wave of adaptive margin strategies. AdaFace~\cite{Adaface} argues that the training strategy should adapt based on image quality, using the feature norm as a proxy to dynamically adjust the margin on a per-sample basis. KappaFace~\cite{KappaFace} tackles the problem from the perspective of class-level difficulty and data imbalance, using the von Mises-Fisher distribution to model the concentration of each class's features and assign a larger margin to harder-to-learn or under-represented classes. ElasticFace~\cite{ElasticFace} proposes that a single fixed margin is too rigid and introduces a randomized margin drawn from a Gaussian distribution, allowing the decision boundary to be extracted and retracted for more flexible learning.

Much of the recent research has also focused on the computational bottleneck of the final fully-connected (FC) layer, whose memory and computational costs scale linearly with the number of identities. Partial FC \cite{PartialFC-2-birds, PartialFC-10-million} maintains the full set of class centers but, for each training step, uses only the positive centers and a randomly sampled subset of the negative centers. This approach dramatically reduces the computational load and memory footprint while maintaining high accuracy, and has been shown to be robust to noise and long-tailed distributions common in web-collected data. 

\subsection{Generalizations of the Cross-Entropy Loss}

Beyond the geometric modifications in face recognition, another line of research has focused on generalizing the cross-entropy loss itself from a probabilistic standpoint. A key development in this area is the framework by Roulet et al. \cite{alpha-loss}, which utilizes Fenchel-Young losses \cite{fenchel-young-losses} to generate a broad family of convex loss functions from f-divergences. This work provides a unified perspective on various losses. For instance, it recovers sparsemax \cite{sparsemax}, originally introduced by Martins and Astudillo as a sparse alternative to $\operatorname{softargmax}$ for attention and multi-label classification, which can produce probability distributions with exact zeros. Chan and Kittler \cite{angularsparsemax} used the sparsemax loss function in conjunction with ArcFace and managed to surpass it using MobileFaceNets~\cite{mobilefacenets}. The concept of sparsemax was further generalized by Peters et al. with entmax~\cite{entmax}, which creates a family of transformations that bridges the gap between dense softmax and sparse sparsemax, offering controllable sparsity for sequence-to-sequence models. The entmax loss is generated by the versatile $\alpha$-divergence. Our work adopts this $\alpha$-divergence loss, but our novelty lies in designing a task-specific reference measure to encode margin penalties.
Another approach to generalization involves using different information-theoretic measures. For instance, Li and Turner proposed using Rényi divergence~\cite{renyi1961measures} for variational inference \cite{li2016renyi}, creating a new family of learning objectives. This work demonstrates how alternate divergences can provide a flexible framework that interpolates between different learning paradigms, moving beyond the direct loss function formulation. Similarly, Novello and Tonello \cite{f-divergence-novello} developed deep learning objectives based on the variational formulation of f-divergences, further highlighting the potential of these constructs to create novel training criteria.
\section{Method}

\subsection{Notation}

Let $k$ be the number of identities in the training set. We denote $[k] = \{1,\dots,k\}$. We denote the probability simplex by $\Delta^{k-1} = \{ \textbf{p}\in\mathbb{R}^{k}_{+}: \langle\textbf{p},\textbf{1}\rangle = 1\}$. We also denote by $\boldsymbol{\theta} \in \mathbb{R}^k$ the output logits produced by a neural network, and $\textbf{y} \in \{e_1,\dots,e_k\}$ denotes the one-hot identity labels. 

Throughout this work, we have adopted the convention proposed by Roulet et al.~\cite{alpha-loss}:
\begin{equation}
    \operatorname{softmax}(\boldsymbol{\theta}) = \log\sum_{{j'}=1}^k \exp(\theta_{j'}),
\end{equation}
\begin{equation}
    [\operatorname{softargmax}(\boldsymbol{\theta})]_j = \frac{\exp(\theta_j)}{\sum_{j'=1}^k\exp(\theta_{j'})}. 
\end{equation}
This convention distinguishes the log-partition function, $\operatorname{softmax}(\boldsymbol{\theta})$ from its gradient, the standard probability-generating $\operatorname{softmax}$ function, $\operatorname{softargmax}(\boldsymbol{\theta})$, as shown in \cref{eq:softmax_grad}. The $\operatorname{softmax}$ is the log-sum-exp operator and $\operatorname{softargmax}$ is the standard $\operatorname{softmax}$ function.
\begin{equation}
    \nabla_{\boldsymbol{\theta}}\operatorname{softmax}(\boldsymbol{\theta}) = \operatorname{softargmax}(\boldsymbol{\theta})
    \label{eq:softmax_grad}
\end{equation}
This equation also holds for the corresponding $\operatorname{softmax}_f$ and $\operatorname{softargmax}_f$ functions discussed below~\cite{alpha-loss}.

\subsection{$\boldsymbol{f}$- and $\boldsymbol{\alpha}$-divergence Loss Framework}

Our methods build upon the generalized framework of Fenchel-Young losses~\cite{fenchel-young-losses}, specifically using the $f$-divergence loss, $l_f$, as proposed by Roulet et al.~\cite{alpha-loss}. 

This framework is derived from a fundamental optimization problem. The generalized operator, $\operatorname{softmax}_f(\boldsymbol{\theta})$ is defined as the solution to a regularized maximization problem:
\begin{equation}
    \operatorname{softmax}_f(\boldsymbol{\theta}) = \max_{\textbf{p}\in\Delta^{k-1}} \langle\textbf{p},\boldsymbol{\theta} \rangle - D_f(\textbf{p}:\textbf{q})
\end{equation}
Here, the f-divergence, $D_f(\textbf{p}:\textbf{q}) = \langle f(\textbf{p}/\textbf{q}), \textbf{q}\rangle$, where $f(\cdot)$ is a convex function, acts as a regularizer. This generalizes the standard $\operatorname{softmax}$ (log-sum-exp) which is recovered when the KL divergence, $D_{KL}(\textbf{p}:\textbf{q})$ is used. The loss function is then defined as:
\begin{equation}
    l_f(\boldsymbol{\theta},\textbf{y};\textbf{q}) = \operatorname{softmax}_f(\boldsymbol{\theta}) + D_f(\textbf{y}:\textbf{q}) - \langle\boldsymbol{\theta}, \textbf{y}\rangle
\end{equation}
where $\boldsymbol{\theta}\in\mathbb{R}^k$ are the logits, $\textbf{y}\in\Delta^{k-1}$ is the ground-truth (in our case a one-hot vector with $1$ at index $y$), $\textbf{q}\in\mathbb{R}^k_+$ is the reference measure (class priors). Note that the term $D_f(\textbf{y}:\textbf{q})$ is constant with respect to the logits $\boldsymbol{\theta}$ and therefore does not influence the gradients. 

The $\alpha$-divergence is a subclass of $f$-divergence, generated by the convex function:
\begin{equation}
    f(u) = \frac{(u^{\alpha} - 1) - \alpha(u-1)}{\alpha(\alpha - 1)}, \quad \alpha\neq 1
    \label{eq:f}
\end{equation}
This framework generalizes the standard Cross-Entropy loss (recovered as $\alpha\to1$ with $\textbf{q}=\textbf{1}$, yielding $f(u) = u\log u -(u-1)$) and allows for controllable sparsity for $\alpha>1$. We refer the reader to \cite{alpha-loss} for the full derivation.

\subsection{Margin-Based Losses}

The standard Cross-Entropy loss, while ensuring separability, proved insufficient to learn the highly discriminative features required for open-set identity recognition. Early approaches like L-Softmax \cite{L-Softmax} and SphereFace~\cite{sphereface} introduced angular margins but faced training issues regarding both numerical stability and convergence. Subsequent research shifted towards additive margins, which proved to be more stable and effective in practice. 

Two prominent examples dominated the field: CosFace~\cite{cosface} and ArcFace~\cite{arcface}. CosFace applies an additive margin directly in the cosine similarity space ($cos(\theta_y) - m$), while ArcFace adds the margin directly to the angle before applying the cosine function $cos(\theta_y + m)$, where $\theta_y \in [0,2\pi)$. Both methods operate on the principle of making the classification task harder for the model. By penalizing the logit (or cosine similarity) $\theta_y$ corresponding to the ground-truth class $y$, they force the model to learn features that are more distinctly separated from other classes, thereby increasing the discriminative power of the learned embeddings and enhancing generalization to identities unseen during training. 

These successful margin-based approaches share a common mechanism: they both directly modify the target logit $\theta_y$ (or its cosine representation) before it is passed to the standard Cross-Entropy loss function. This geometric manipulation of the logits contrasts with the approach taken by Q-Margin, which achieves a similar discriminative effect by modifying the reference measure $\textbf{q}$ within the $\alpha$-divergence framework.

\subsection{Our Methods}

Building on the generalized $\alpha$-divergence loss framework and inspired by the efficacy of margin penalties in face recognition, we propose two different methods. Observing that a margin penalty, such as the one in CosFace, can be incorporated either by modifying the reference measure $\textbf{q}$ or by directly modifying the (unnormalized) log-likelihood of the ground-truth label (logit $\theta_y$), we explore both ways and propose two different methods.

\subsubsection{Q-Margin: Modifying the Reference Measure}

Our first contribution, Q-Margin, introduces the margin penalty through the reference measure $\textbf{q}$. Instead of directly manipulating the scaled cosine similarities $\boldsymbol{\theta} = s\cdot \textbf{c}$, where $\textbf{c}\in [-1,1]^k$ is the vector of cosine similarities between a sample's embedding and the $k$ class prototype vectors, we encode the margin penalty into $\textbf{q}$. For a sample with ground-truth label $y$, the reference measure is defined as:

\begin{equation}
\label{eq:prior}
    q_j = \left \{
    \begin{array}{ll}
         \exp{(-s\cdot m)}&  j = y\\
         1 & j \neq y 
    \end{array}
    \right.,
\end{equation}
where $s$ is the scaling factor and $m$ is the margin hyperparameter. By significantly down-weighting $q_y$, we create a stricter learning objective. Minimizing the $\alpha$-divergence loss $l_f(\boldsymbol{\theta},\textbf{y};\textbf{q})$ compels the model to produce a substantially higher target logit $\theta_y$ relative to non-target logits to compensate, implicitly creating the desired decision margin through probabilistic means rather than direct geometric logit modification. The final loss is
\begin{equation}
    L_{Q-Margin} = l_f(s \cdot \textbf{c}, \textbf{y}; \textbf{q}).
\end{equation}
It is easy to show that CosFace can be considered as a special case of Q-Margin, by setting $\alpha=1$.

\subsubsection{Gradients and Posterior Sparsity} \label{sec:grads}

The gradients for the $\alpha$-divergence loss, $l_f$, are key to understanding the model's behavior. The gradient w.r.t. the $j$-th logit $\theta_j$ (assuming target identity $y$) is $\frac{\partial l_f}{\partial\theta_j}=[\operatorname{softargmax}_f(\boldsymbol{\theta})]_j - {\bf y}_j$. The $j$-th element of the $\alpha$-$\operatorname{softargmax}$, which we denote as the posterior probability $p_j$, is given by: 

\begin{equation}
\begin{split}
    p_j & = [\operatorname{softargmax}_f(\boldsymbol{\theta})]_j  \\
    & = q_{j}[1+({\alpha-1})(\theta_j - \tau^*)]_{+}^{1/(\alpha-1)}
\end{split}
\label{eq:l_grad_theta}
\end{equation}
where $\tau^*$ is a scalar value, computed iteratively, that ensures the probabilities sum to one $\left(\sum_jp_j = 1\right)$. For $\alpha>1$, the $[\cdot]_+ = \max(0,\cdot)$ operator is active, meaning that if a logit $\theta_j$ falls below a certain threshold relative to $\tau^*$, the posterior probability $p_j$ becomes exactly 0, resulting in sparse posterior distributions.

It is instructive to note the relationship between this formulation and the standard $\operatorname{softargmax}$. As $\alpha\to1$, the $p_j$ term recovers the standard exponential function $\exp(x) = \lim_{n\to \infty}\left(1+\frac{x}{n}\right)^n$. By setting $x=\theta_j-\tau^*$ and $n=1/(\alpha-1)$, we see that $\lim_{\alpha \to 1}p_j = \exp{(\theta_j - \tau^* + \log q_j)}$. This recovers the form of the standard $\operatorname{softargmax}$ probability, making the $\alpha$-divergence loss a polynomial generalization of the standard cross-entropy loss, and the CosFace loss when $\textbf{q}$ is as in \cref{eq:prior}.

\subsubsection{Efficient Computation of $\boldsymbol{\alpha}$-$\boldsymbol{\operatorname{softargmax}}$}

The $\alpha$-$\operatorname{softargmax}$ (\cref{eq:l_grad_theta}) is computed by first finding the scalar $\tau^*$. This is the unique solution for $\tau$ in the root finding equation:
\begin{equation}
    \sum_{j=1}^k q_jf_*'(\max\{\theta_j-\tau, f'(0)\}) = 1.
\end{equation}
Here, $f$ is the convex function generating the $\alpha$-divergence (\cref{eq:f}), $f'(u) = \frac{u^{\alpha-1}-1}{\alpha-1}$, $\alpha \neq 1$, the derivative of $f$, $f^*(v) = \frac{1}{\alpha}([1+(\alpha-1)v]^{\frac{\alpha}{\alpha-1}}_+ - 1)$, $\alpha\in(1,+\infty)$ is the convex conjugate of $f$ (i.e., $\operatorname{softmax}_f$), $f_*'=(f^*)'$ the derivative of the convex conjugate (i.e., $\operatorname{softargmax}_f$). The components $q_j$ are from the reference measure and $\theta_j$ are the input logits. 

Following Roulet et al.~\cite{alpha-loss}, this value $\tau^*$ is found efficiently with the iterative bisection algorithm. The search is bounded within the range $[\tau_{\min},\tau_{\max}]$ where for $t \in \operatorname{argmax}_{j \in [k]} \theta_j$ it is \ $\tau_{\min} = \theta_{t} - f'(1 / q_{t})$ and $\tau_{\max} = \theta_{t} - f'\left(1 / \sum\nolimits_{j} q_j\right)$.

\subsubsection{Alpha-Additive-Angular Margin (\textbf{\methodB}) Loss}

An alternative approach, Alpha-Additive-Angular Margin (or \methodB) Loss, applies the margin penalty directly to the logits, analogous to other margin-based methods. The motivation for this method stems from the sparsity-inducing gradients of the $\alpha$-divergence loss detailed in \Cref{sec:grads}. Empirically, we observed that the zero-forcing property ($p_j=0$ if $\theta_j$ is below a certain threshold) also affected the target class, particularly during the initial training phases when the classification weights ($\textbf{W}$) might be suboptimal. Our analysis revealed that for a significant number of training examples, the posterior probability assigned to the ground-truth class became exactly zero ($p_y=0$).

\subsubsection{A3M-I: Stabilizing A3M with Prototype Re-Initialization}
\label{sec:a3m-i}
To address the optimization instability detailed in \Cref{sec:grads}, our method \methodB-I enhances the \methodB\ loss by incorporating a mid-training prototype weight initialization step. This procedure aims to replace potentially suboptimal prototypes, which, due to the sparsity-inducing nature of the loss, have been updated very rarely and are not well aligned with the embedding extractor, with more optimal ones derived directly from the feature space. 

The base \methodB\ (Alpha-Additive-Angular Margin) loss used in this procedure applies the standard ArcFace margin geometrically to the logits \emph{before} they are passed to the $\alpha$-divergence loss (with a uniform $\textbf{q}=\textbf{1}$). It is defined as:
\begin{equation}
    L_{\methodB} = l_f(s\cdot \textbf{c}',\textbf{y};\textbf{q}=\textbf{1})
\end{equation}
where $\textbf{c}'$ represents the margin-modified cosine similarities. 

The \methodB-I procedure begins by training the model for a few epochs (3 and 20 in face recognition and speaker verification, respectively) using this $L_{\methodB}$ loss. Following this initial phase, training is paused to perform a pseudo-epoch over the training data using the backbone in evaluation mode. During this pass, the sum of feature embeddings for each identity is accumulated and then $L_2$-normalized to compute the identity prototypes $\textbf{W}_{proto}$. The weights $\textbf{W}$ of the final classification head are then replaced by the newly computed prototypes $\textbf{W}_{proto}$ and the corresponding optimizer states are reset. Training is then resumed using the same $L_{\methodB}$ loss function. By providing these data-derived and potentially more optimal prototypes, we address the issue of estimating zero probability of the target identity and stabilize training.
\section{Face Recognition Experiments}

\subsection{Training and Testing Data}

Our models were trained on two large-scale public datasets: MS1MV3 \cite{ms1mv3}, a refined version of the MS-Celeb-1M dataset, containing approximately 5.8 million images of 93K identities, and WebFace42M \cite{webface260m}, a comprehensively cleaned subset of the WebFace260M dataset containing 42 million images of 2 million identities. Following standard preprocessing protocols in face recognition, all images were pre-processed using alignment and resizing to a final resolution of 112 $\times$ 112 pixels.

During training, several standard verification benchmarks were used as development sets to monitor performance. These datasets are: LFW \cite{lfw}, CFP-FP \cite{cfp-fp}, AgeDB-30 \cite{agedb}, CALFW \cite{ca-lfw} and CPLFW \cite{cp-lfw}. For these datasets, we report the False Rejection Rate (FRR) at a False Acceptance Rate (FAR) of $10^{-3}$. Note that FRR = 1 - TAR (True Acceptance Rate).

The final evaluation was performed on the more challenging IJB-B \cite{ijb-b} and IJB-C \cite{ijb-c} benchmarks. These datasets are designed to test recognition performance in unconstrained scenarios. On these two test sets, we report FRR at FAR levels of $10^{-4}$ and $10^{-5}$ as our primary performance metrics, which represent standard operating points for evaluating these benchmarks.

All experiments with the MS1MV3 dataset were run three times, and we report the FRR averaged across the runs. Models trained with WebFace42M, however, were trained a single time due to the significant computational cost.

\subsection{Implementation Details}

The experimental framework was built upon the InsightFace toolkit\footnote{\url{https://github.com/deepinsight/insightface}}, specifically its ArcFace implementation, with minimal modifications to integrate the custom $\alpha$-divergence loss function. Training was carried out with 8 NVIDIA A100 (40GB) GPUs.

We evaluated two standard backbone architectures from the ResNet family \cite{resnet}: ResNet-50 (R-50) and ResNet-100 (R-100). All models were trained for 20 epochs using Stochastic Gradient Descent (SGD) as an optimizer. We employed a step-wise learning rate decay schedule, starting at 0.1 for the first 7 epochs, decreasing to 0.01 for the next 6 epochs, and concluding at 0.001 for the final 7 epochs. A momentum of 0.9 and a weight decay of $5\times10^{-4}$ were used. The batch size was set to 128 per GPU, resulting in an effective total batch size of 1024.

The \methodB-I method applied the mid-training prototype re-initialization technique detailed in \Cref{sec:a3m-i}. This procedure was performed after the third epoch and involved computing $L_2$-normalized prototypes from the summed feature embeddings to replace the classification head weights.

\subsection{Results}
\label{sec:results}

We first provide a quantitative analysis in \Cref{tab:sparsity} of the misalignment between prototypes and embeddings (measured by identities and examples for which $p_y=0$) and the resulting posterior sparsity. The $\methodB$ loss, which combines a geometric margin $m=0.5$ with $\alpha=1.25$, suffers from severe misalignment: 24.12\% of all training images produced a zero probability for the ground-truth class, and 5.67\% of all identities had every image of the identity produce a zero probability for the ground-truth class. Our proposed methods virtually eliminate this problem. \methodB-I (also with $m=0.5$) reduces the image-level misalignment to 1.52\% while Q-Margin's probabilistic margin almost completely solves it (0.15\%). 

Crucially, this stability is achieved while preserving the primary benefit of $\alpha$-divergence loss, as all methods produce an output vector that is $>99.97\%$ sparse. Ideally, this extreme sparsity allows for the construction of a memory-efficient training framework where logit evaluation is constrained to neighboring prototypes. Such an approach, facilitated by sparse embedding layers and optimizers, would yield significant computational and memory gains without resorting to approximate methods.

The final column reveals another important behavioral difference between the methods. \methodB-I still produces \emph{hard} one-hot predictions for 27.44\% of images. In contrast, Q-Margin does so for only 0.14\% of images. This suggests that Q-Margin's probabilistic margin acts as a softer regularizer that prevents over-confidence, whereas \methodB-I retains the \emph{hard} decision-making characteristics of the base \methodB\ method.

\begin{table}[h!]
    \centering
    \addtolength{\tabcolsep}{-2pt} 
    \caption{Statistics of sparsity and prototype misalignment of R-50 models trained on MS1MV3. We compare misalignment of prototypes (using $p_y=0$ as a cue) against the overall posterior sparsity, demonstrating that our methods address the misalignment while retaining the desired sparsity of $\alpha$-divergence. The statistics are collected after the final epoch.}
    \label{tab:sparsity}
    \resizebox{1.0\linewidth}{!}{
    \begin{tabular}{l | cc | cc}
        \toprule
         & \multicolumn{2}{c|}{Misalignment $p_y=0$ (\%)} & \multicolumn{2}{c}{Sparsity Statistics (\%)} \\
        \cmidrule(r){2-3} \cmidrule(l){4-5}
        Loss & \# Ids & \# Images & \parbox{2.8cm}{\centering Posterior \\ Sparsity} & \parbox{2.8cm}{\centering Images with 1 \\ Non-Zero $p_j$} \\
        \midrule
        A3M     & 5.67 & 24.12 & 99.994  & 27.58  \\
        A3M-I   & 0.15 & 1.52  & 99.994  & 27.44  \\
        Q-Margin & 0.03 & 0.15  & 99.976  & 0.14 \\
        \bottomrule
    \end{tabular}}
\end{table}

We present a comprehensive performance evaluation of the proposed Q-Margin and \methodB-I methods against the strong margin-based losses ArcFace \cite{arcface} and CosFace \cite{cosface}. As shown in \Cref{tab:results-face}, both proposed methods consistently match or outperform the baselines across all experimental setups, with the most significant improvements observed on the challenging IJB-B and IJB-C benchmarks.

\begin{table*}[h!]
\addtolength{\tabcolsep}{-3pt}
\caption{{Results for both ResNet-50 (R-50) and ResNet-100 (R-100) models which were trained with the MS1MV3 dataset and the WebFace dataset respectively. For all datasets, we report the False Rejection Rate (FRR, \%) at a given False Acceptance Rate (FAR). For the Benchmark datasets, we report results for only one operating point, FAR=$10^{-3}$, and for the IJB datasets at the $10^{-4}$ and $10^{-5}$ operating points. \methodB-I and Q-Margin-I are configurations of \methodB\ and Q-Margin respectively with Prototype Re-Initialization.}}
\resizebox{1.0\linewidth}{!}{
\begin{tabular}{c|c|ccc|ccccc|cc|cc|cc|cc} 
        \toprule
         & & \multicolumn{3}{c|}{Parameters} & \multicolumn{5}{c|}{Benchmarks} & \multicolumn{2}{c|}{IJB-B} & \multicolumn{2}{c|}{IJB-C} & \multicolumn{2}{c|}{Avg. Impr. IJB-B (\%)} & \multicolumn{2}{c}{Avg. Impr. IJB-C (\%)}\\
         \cmidrule(lr){3-5} \cmidrule(lr){6-10} \cmidrule(lr){11-12} \cmidrule(lr){13-14} \cmidrule(lr){15-16} \cmidrule(l){17-18}
        Loss & Model & $\alpha$ & s & m & LFW & CFP-FP & AgeDB-30 & CALFW & CPLFW & $10^{-4}$ & $10^{-5}$ & $10^{-4}$ & $10^{-5}$ & ArcFace & CosFace & ArcFace & CosFace \\
        \midrule
ArcFace & R-50 & - & 65 & 0.5 & 0.17 & 1.06 & 1.76 & 3.89 & 6.90 & 5.31 & 10.30 & 3.82 & 5.97 & - & -1.07 & - & -0.69\\
CosFace & R-50 & - & 64 & 0.5 & 0.17 & 1.12 & 1.72 & 3.93 & 6.97 & 5.30 & \textbf{9.94} & 3.80 & 5.97 & 1.06 & - & 0.69 & -\\
\methodB & R-50 & 1.25 & 64 & 0.5 & 0.18 & 1.10 & 1.79 & 3.81 & 7.06 & 5.21 & 10.13 & \textbf{3.75} & 5.90 & 1.39 & 0.33 & 1.22 & 0.54\\
\methodB & R-50 & 2 & 1.9 & 0.2 & 0.18 & 1.03 & 1.72 & 3.93 & 6.89 & 5.49 & 10.97 & 3.99 & 6.28 & -3.06 & -4.17 & -4.69 & -5.41\\
Q-Margin-I & R-50 & 1.25 & 32 & 0.2 & \textbf{0.16} & \textbf{1.02} & 1.82 & \textbf{3.78} & \textbf{6.82} & 5.24 & 10.39 & 3.77 & 5.87 & 2.10 & 1.05 & 1.38 & 0.70\\
\methodB-I & R-50 & 1.25 & 64 & 0.5 & 0.18 & 1.17 & 1.76 & 3.87 & 7.13 & 5.33 & 10.18 & 3.83 & 5.95 & 1.05 & -0.01 & 0.10 & -0.59\\
\methodB-I & R-50 & 2 & 1.9 & 0.2 & 0.17 & 1.03 & 1.69 & 3.88 & 6.83 & 5.38 & 10.77 & 3.92 & 6.26 & -1.41 & -2.50 & -2.94 & -3.65\\
Q-Margin & R-50 & 1.5 & 20 & 0.2 & 0.17 & 1.09 & \textbf{1.68} & 3.87 & 7.01 & 5.30 & 10.18 & 3.81 & 6.09 & 0.75 & -0.31 & -0.63 & -1.32 \\
\rowcolor[HTML]{C0C0C0}{Q-Margin} & R-50 & 1.25 & 32 & 0.2 & 0.20 & 1.07 & 1.82 & 3.86 & 6.91 & \textbf{5.15} & 9.96 & \textbf{3.75} & \textbf{5.76} & \textbf{3.87} & \textbf{2.84} & \textbf{2.65} & \textbf{1.97}\\
\midrule
ArcFace & R-100 & - & 64 & 0.5 & 0.15 & \textbf{0.64} & 1.57 & \textbf{3.82} & 5.27 & 3.91 & 7.74 & 2.46 & 4.30 & - & -0.94 & - & 1.42\\
CosFace & R-100 & - & 64 & 0.5 & 0.18 & 0.80 & 1.72 & 3.92 & 5.48 & 3.95 & 6.85 & 2.51 & 4.18 & 0.83 & - & -1.46 & -\\
\methodB & R-100 & 1.25 & 64 & 0.5 & 0.17 & 0.66 & 1.87 & 4.07 & 5.23 & 3.87 & 6.94 & 2.45 & 3.89 & 2.24 & 1.38 & 3.31 & 4.70\\
\methodB & R-100 & 2 & 1.9 & 0.2 & 0.18 & 0.87 & 2.02 & 3.93 & 5.92 & 4.50 & 8.16 & 2.83 & 4.72 & -17.78 & -18.80 & -17.12 & -15.44\\
Q-Margin-I & R-100 & 1.25 & 32 & 0.2 & 0.17 & \textbf{0.64} & 1.72 & 3.88 & 5.43 & 3.84 & 7.08 & 2.38 & 3.96 & 3.00 & 2.14 & 3.43 & 4.82\\
\methodB-I & R-100 & 1.25 & 64 & 0.5 & 0.18 & 0.76 & 1.83 & 3.93 & 5.38 & 3.71 & \textbf{6.47} & 2.34 & \textbf{3.47} & \textbf{7.01} & \textbf{6.23} & 7.67 & 9.03\\
\methodB-I & R-100 & 2 & 1.9 & 0.2 & 0.18 & 0.93 & 1.97 & 3.95 & 6.38 & 4.61 & 8.38 & 3.01 & 4.91 & -19.25 & -20.30 & -22.53 & -20.77\\
Q-Margin & R-100 & 1.5 & 32 & 0.2 & 0.17 & 0.64 & 1.57 & \textbf{3.82} & 5.27 & 3.88 & 7.33 & 2.50 & 4.51 & -0.05 & -0.98 & -2.47 & -1.03 \\
\rowcolor[HTML]{C0C0C0}{Q-Margin} & R-100 & 1.25 & 32 & 0.2 & \textbf{0.13} & \textbf{0.64} & \textbf{1.43} & 3.85 & \textbf{5.17} & \textbf{3.70} & 6.62 & \textbf{2.24} & 3.65 & 5.55 & 4.73 & \textbf{9.22} & \textbf{10.53}\\
\bottomrule
    \end{tabular}}
\label{tab:results-face}
\end{table*}

The advantage is particularly evident on the large-scale WebFace42M dataset. Here, our methods exhibit a compelling trade-off at different operating points. Q-Margin ($\alpha=1.25$, $s=32$, $m=0.2$) achieves the best performance at the FRR@FAR=$10^{-4}$ threshold reducing the error on IJB-C to 2.24\% (vs. 2.46\% for ArcFace). This configuration yields the highest average improvement on IJB-C (9.22\% and 10.53\% over ArcFace and CosFace, respectively). At the strictest FRR@FAR=$10^{-5}$ threshold, \methodB-I ($\alpha=1.25$, $s=64$, $m=0.5$) achieves its optimal performance, delivering the lowest error on both IJB-B and IJB-C and providing the best average improvement on IJB-B (7.01\% over ArcFace and 6.23\% over CosFace).

\begin{figure}[h!]
        \centering
        \setlength{\abovecaptionskip}{2pt}
        \includegraphics[width=0.48\textwidth]{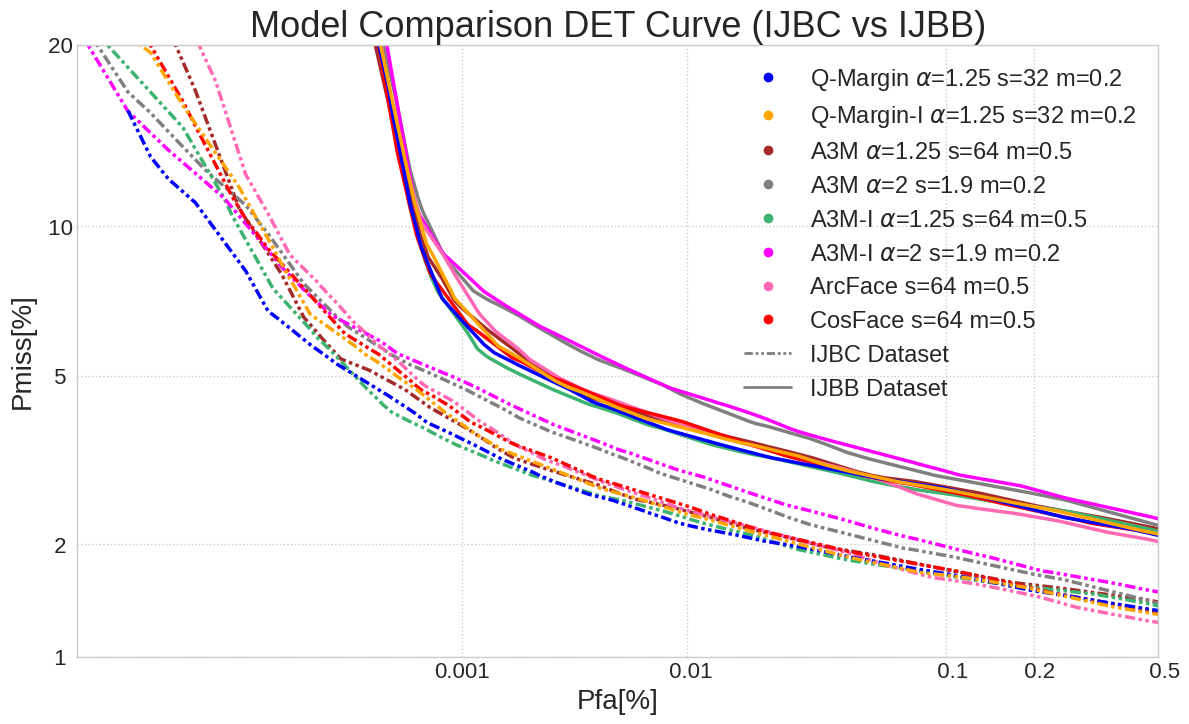} 
        \caption{False acceptance vs missed detection probabilities for various losses on IJB-B and IJB-C, focused on low false acceptance operating points.\methodB-I and Q-Margin-I are configurations of \methodB\ and Q-Margin, respectively, with Prototype Re-Initialization.}
        \label{fig:det-curve}
        \vspace{-16pt}
\end{figure}

The Detection Error Tradeoff (DET) curves in \Cref{fig:det-curve} provide a visual corroboration of these results. Our best Q-Margin configuration (blue line) is shown to consistently outperform the ArcFace (pink) and CosFace (red) baselines. The plot also clearly illustrates the trade-off between our two proposed methods: Q-Margin excels around $10^{-4}$ FAR, while the prototype re-initializing \methodB-I (green) demonstrates superior performance at the stricter $10^{-5}$ FAR, reinforcing our quantitative results from \Cref{tab:results-face}.

\begin{figure*}[h!]
    \centering
    \includegraphics[width=\textwidth]{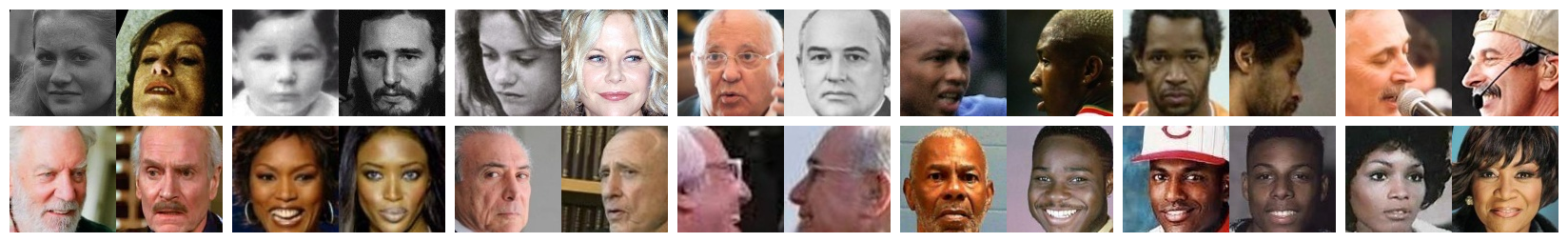}
    \caption{Challenging pairs from AgeDB-30, CALFW, and CPLFW where our proposed method (Q-Margin, $\alpha$=1.25, s=32, m=0.2) succeeds, while the ArcFace baseline fails. The top row shows genuine pairs correctly accepted by our model and falsely rejected by the baseline. The bottom row shows impostor pairs correctly rejected by our model and falsely accepted by the baseline.}
    \label{fig:clean_grid}
    \vspace{-16pt}
\end{figure*}

Beyond quantitative metrics, \Cref{fig:clean_grid} provides qualitative evidence of Q-Margin's improved robustness. The examples show our method succeeding where ArcFace fails, correctly verifying genuine pairs despite large intra-class variations (\eg extreme pose, aging) and correctly rejecting impostor pairs. This suggests Q-Margin learns a more robust and discriminative embedding space.

\subsubsection{Ablation}

To isolate the contribution of each component, we conducted two main ablations, detailed in \Cref{tab:results-face}. To analyze the impact of \methodB-I's mid-training prototype re-initialization, we compared its performance against the base \methodB. The results show this technique is critically scale-dependent: on the smaller MS1MV3 dataset, re-initialization offered no clear advantage. However, this trend reversed on the massive WebFace42M dataset. Here, the \methodB\ method still outperformed ArcFace and CosFace, but the full \methodB-I method, with re-initialization, proved to be a critical component for top performance. This validates our hypothesis that this technique is helpful in tackling the optimization challenges of large-scale noisy data.
To complete this study, we also applied the re-initialization technique to Q-Margin (termed Q-Margin-I in \Cref{tab:results-face}). This failed to improve performance on the IJB datasets, suggesting that the probabilistic margin of Q-Margin does not suffer from the same instability associated with the geometric margin of \methodB.

\section{Speaker Verification Experiments}

\subsection{Training and Testing Data}

For speaker verification, we adopted the standard protocol established by the VoxCeleb benchmarks. The training corpus is the development set of VoxCeleb2~\cite{Voxceleb2}, comprising about 1M utterances from 5,994 speakers.

For evaluation, we adopted the challenging trial list termed VoxCeleb1-H (hard). It contains 550,894 trials featuring 1,190 speakers. The difficulty of the evaluation set arises from the constraint that trials consist of utterances from speakers of the same gender and nationality. While other trial lists (VoxCeleb1-O and VoxCeleb1-E) exist, we focus on this \emph{hard} set to maintain alignment with the challenging scenarios evaluated in the face recognition experiments. While experiments on the other trial lists yielded similar conclusions, we omit those results for brevity. Our analysis focuses on FRR(\%) at FAR of $10^{-3}$ and $10^{-4}$.

\begin{table}[t]
    \centering
    \addtolength{\tabcolsep}{-3pt}
    \small
    \caption{Results with a ResNet-34 (R-34) model trained on VoxCeleb2 dev. We report the False Rejection Rate (FRR, \%) at False Acceptance Rates (FAR) of $10^{-3}$ and $10^{-4}$. \methodB-I applies identity prototypes' re-initialization. The experiments are performed using VoxCeleb1-H as test set.}
    \label{tab:results-r34-vox2dev} 
    
    \begin{tabular}{l| ccc| cc| cc} 
        \toprule
         & \multicolumn{3}{c|}{Parameters} &  \multicolumn{2}{c|}{FRR@FAR} & \multicolumn{2}{c}{Avg. Impr. (\%)} \\
        \cmidrule(lr){2-4} \cmidrule(lr){5-6} \cmidrule(l){7-8} 
        Loss & $\alpha$ & s & m & $10^{-3}$ & $10^{-4}$ & ArcFace & CosFace \\
        \midrule
        ArcFace & - & 32 & 0.2 & 13.38 & 27.92 & - & -2.64 \\
        CosFace & - & 32 & 0.2 & 13.31 & 29.39 & 2.48 & - \\
        \midrule
        \methodB & 2.0 & 5 & 0.1 & 12.59 & 25.63 & 8.16 & 5.75 \\
        \methodB & 1.75 & 5 & 0.1 & 12.31 & 25.95 & 10.06 & 7.71 \\
        \methodB-I & 2.0 & 5 & 0.1 & 12.68 & 26.41 & 9.07 & 6.70\\
        \methodB-I  & 1.75 & 5 & 0.1 & 12.17 & 25.66 & 10.79 & 8.45 \\
        Q-Margin & 1.75 & 5  & 0.1 & 12.98 & \textbf{25.40} & 7.69 & 5.25 \\
        \rowcolor[HTML]{C0C0C0}{Q-Margin}& 1.5  & 10 & 0.1 & \textbf{12.09} & 27.04 & \textbf{11.78} & \textbf{9.49} \\
        \bottomrule
    \end{tabular}
    \vspace{-16pt}
\end{table}
\subsection{Implementation Details}

Our implementation is based on the WeSpeaker toolkit\footnote{\url{https://github.com/wenet-e2e/wespeaker}}~\cite{wang2024advancing}. Training was conducted on 4 NVIDIA RTX A4000 (16GB) GPUs. A ResNet-34 (R-34) backbone was employed as the feature extraction network. The model was trained for 150 epochs using SGD with a momentum of 0.9 and a weight decay of $10^{-4}$. The learning rate followed a two-phase schedule: a linear warm-up during the first six epochs, after which it decreased exponentially from 0.1 to $5 \cdot 10^{-5}$ over the remaining training period. Following common practice in speaker verification, the margin parameter $m$ was annealed, starting from 0.0 and increasing exponentially between epochs 20 and 40 to its target value; this value was subsequently held constant. A batch size of 128 per GPU was used. To improve robustness, input data was augmented with additive noise and reverberation during training.

\subsection{Results}

Consistent with the face recognition evaluation, we compare our proposed methods with both CosFace and ArcFace baselines. The overall results are presented in \Cref{tab:results-r34-vox2dev}, showing FRR for different FAR values and relative improvements over the baselines on the VoxCeleb1-H evaluation dataset of interest. The results demonstrate the strong performance of our methods and their sensitivity to hyperparameters ($\alpha$, $s$, $m$). For a given $\alpha$, optimal values for scale $s$ and margin $m$ were estimated via empirical search from the sets $s \in \{5, 10, 20, 32\}$ and $m \in \{0.1, 0.2, 0.3\}$. 

In contrast to the face recognition results (which favored $\alpha=1.25$), we observed that higher values of $\alpha$ ($1.5$ and $1.75$) lead to better-performing speaker verification systems. We hypothesize that this phenomenon is related to the significantly smaller number of training identities. Conversely, these results reinforce observations made in \Cref{tab:results-face}, i.e., that the most consistent improvements are achieved with moderate values of scale ($s\in\{5,10\}$) and margin ($m=0.1$). Specifically, we observed an interdependence of $\alpha$ and $s$, where a higher $\alpha$ requires a smaller $s$ and vice versa. We hypothesize this is related to the achievable peakiness of $\operatorname{softargmax}_f$ function (which directly affects optimization via the gradients in \cref{eq:l_grad_theta}.

\begin{figure}[h!]
    \centering
    \includegraphics[width=\linewidth,trim={1cm 0.3cm 1cm 0.4cm},clip]{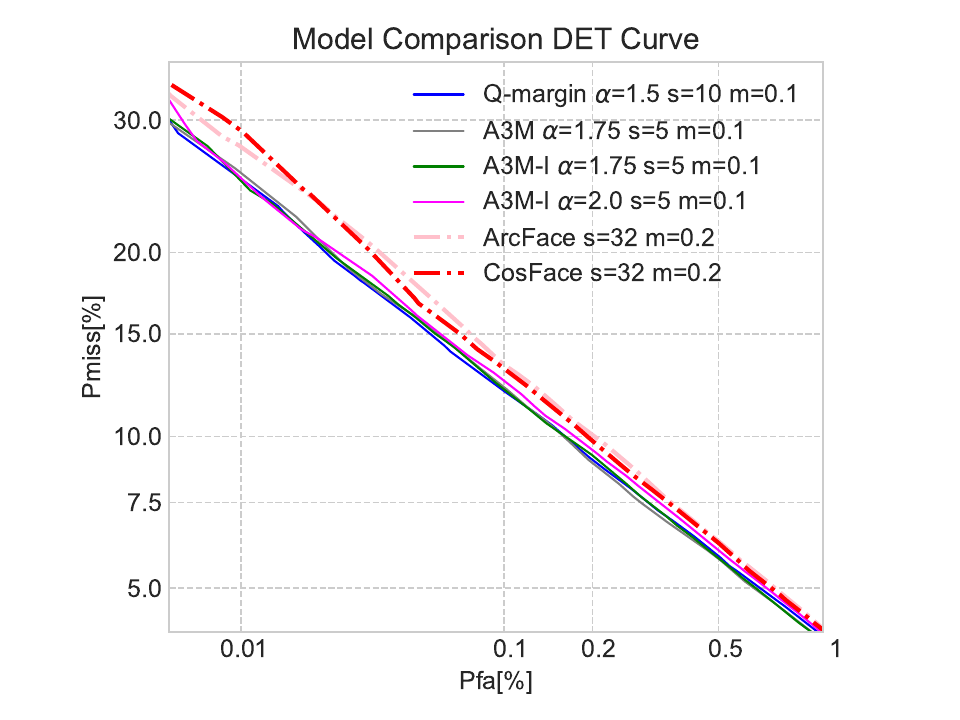} 
    \caption{False acceptance vs missed detection probabilities for various losses on VoxCeleb1-H, focused on low false acceptance operating points. \methodB-I is a configuration of $\methodB$ with Prototype re-initialization.}
    \label{fig:det-curve-speaker}
\end{figure}

Both Q-Margin and \methodB\ yield strong results at the two operating points of interest. To provide a broader assessment, we also report the average relative improvement over the interval between $10^{-3}$ and $10^{-4}$ FAR. We observe a significant improvement from Q-Margin ($\alpha=1.5, s=10, m=0.1$) over both baselines. Its performance is closely followed by that of A3M-I ($\alpha = 1.75, s = 5, m = 0.1$).

While not obvious from the discrete operating points, the averaged improvements demonstrate the utility of the prototype re-initialization used in \methodB-I. This is evident when comparing the \methodB-I and baseline \methodB\ results in the table. To visually illustrate these outcomes, we present DET curves in the low-FAR region for selected systems in \Cref{fig:det-curve-speaker}. While the curves for CosFace and ArcFace nearly overlap, the curves for models using our proposed losses are visibly separated, indicating superior performance.

\section{Conclusions}
Margin-based loss functions have driven the success of embedding-based networks in face recognition and speaker verification, while recent work on generalized cross-entropy ($\alpha$-divergence losses) has enabled principled control of sparsity in probability distributions and class-dependent penalization through the reference measure \textbf{q}. We propose a synergistic combination of these two paradigms and resolve a core challenge: sparsity-inducing gradients interact destructively with geometric margins, causing misaligment between prototypes and embeddings. Our method bridges information-theoretic loss design with geometric margin techniques.

Our first approach, Q-Margin, introduces a \textit{probabilistic margin} by encoding margin penalties directly into a non-uniform \textbf{q}, avoiding gradient instability and offering strong robustness on large, noisy datasets. Our second method, \methodB-I, provides an \textit{optimization-level} remedy that preserves the standard geometric margin but mitigates misalignment via mid-training prototype re-initialization using data-derived representations.

We targeted low-FAR operating regimes and show that both methods consistently outperform strong ArcFace and CosFace baselines on challenging face (IJB-B, IJB-C) and speaker (VoxCeleb1-H) benchmarks. Ablations also reveal practical hyperparameter settings and show that optimal $\alpha$ values are modality-dependent.

While A3M-I currently relies on geometric margins and prototype re-initialization, future work will explore theoretical formulations that avoid explicit geometric margins and integration with methods that adaptively adjust margins. In our framework, it would correspond to dynamically updating the non-uniform reference measure. Finally, we plan to exploit the inherent sparsity of the proposed losses towards a memory-efficient implementation.   
{
    \bibliographystyle{ieeenat_fullname}
    \bibliography{main}

@String(CVPR= {IEEE Conf. Comput. Vis. Pattern Recog.})

@String(ICPR = {Int. Conf. Pattern Recog.})

@String(ICASSP=	{ICASSP})

@String(CVPRW= {IEEE Conf. Comput. Vis. Pattern Recog. Worksh.})

@String(CVPR  = {CVPR})

@String(ICPR  = {ICPR})

@String(CVPRW= {CVPRW})

@INPROCEEDINGS{arcface,
  author={Deng, Jiankang and Guo, Jia and Xue, Niannan and Zafeiriou, Stefanos},
  booktitle={2019 IEEE/CVF Conference on Computer Vision and Pattern Recognition (CVPR)}, 
  title={ArcFace: Additive Angular Margin Loss for Deep Face Recognition}, 
  year={2019},
  volume={},
  number={},
  pages={4685-4694},
  doi={10.1109/CVPR.2019.00482}}

@INPROCEEDINGS{cosface,
  author={Wang, Hao and Wang, Yitong and Zhou, Zheng and Ji, Xing and Gong, Dihong and Zhou, Jingchao and Li, Zhifeng and Liu, Wei},
  booktitle={2018 IEEE/CVF Conference on Computer Vision and Pattern Recognition}, 
  title={CosFace: Large Margin Cosine Loss for Deep Face Recognition}, 
  year={2018},
  volume={},
  number={},
  pages={5265-5274},
  doi={10.1109/CVPR.2018.00552}}

@INPROCEEDINGS{sphereface,
  author={Liu, Weiyang and Wen, Yandong and Yu, Zhiding and Li, Ming and Raj, Bhiksha and Song, Le},
  booktitle={2017 IEEE Conference on Computer Vision and Pattern Recognition (CVPR)}, 
  title={SphereFace: Deep Hypersphere Embedding for Face Recognition}, 
  year={2017},
  volume={},
  number={},
  pages={6738-6746},
  doi={10.1109/CVPR.2017.713}}

@InProceedings{sparsemax,
  title = 	 {From Softmax to Sparsemax: A Sparse Model of Attention and Multi-Label Classification},
  author = 	 {Martins, Andre and Astudillo, Ramon},
  booktitle = 	 {Proceedings of The 33rd International Conference on Machine Learning},
  pages = 	 {1614--1623},
  year = 	 {2016},
  editor = 	 {Balcan, Maria Florina and Weinberger, Kilian Q.},
  volume = 	 {48},
  series = 	 {Proceedings of Machine Learning Research},
  address = 	 {New York, New York, USA},
  month = 	 {20--22 Jun},
  publisher =    {PMLR},
  url = 	 {https://proceedings.mlr.press/v48/martins16.html},
}

@inproceedings{entmax,
    title = "Sparse Sequence-to-Sequence Models",
    author = "Peters, Ben  and
      Niculae, Vlad  and
      Martins, Andr{\'e} F. T.",
    editor = "Korhonen, Anna  and
      Traum, David  and
      M{\`a}rquez, Llu{\'i}s",
    booktitle = "Proceedings of the 57th Annual Meeting of the Association for Computational Linguistics",
    month = jul,
    year = "2019",
    address = "Florence, Italy",
    publisher = "Association for Computational Linguistics",
    url = "https://aclanthology.org/P19-1146/",
    doi = "10.18653/v1/P19-1146",
    pages = "1504--1519",
}

@INPROCEEDINGS{Adaface,
  author={Kim, Minchul and Jain, Anil K. and Liu, Xiaoming},
  booktitle={2022 IEEE/CVF Conference on Computer Vision and Pattern Recognition (CVPR)}, 
  title={AdaFace: Quality Adaptive Margin for Face Recognition}, 
  year={2022},
  volume={},
  number={},
  pages={18729-18738},
  doi={10.1109/CVPR52688.2022.01819}}

@inproceedings{alpha-loss,
title={Loss Functions and Operators Generated by f-Divergences},
author={Vincent Roulet and Tianlin Liu and Nino Vieillard and Michael Eli Sander and Mathieu Blondel},
booktitle={Forty-second International Conference on Machine Learning},
year={2025},
url={https://openreview.net/forum?id=V1YfPJDliw}
}

@INPROCEEDINGS {VGGFace2,
author = { Cao, Qiong and Shen, Li and Xie, Weidi and Parkhi, Omkar M. and Zisserman, Andrew },
booktitle = { 2018 13th IEEE International Conference on Automatic Face \& Gesture Recognition (FG 2018) },
title = {{ VGGFace2: A Dataset for Recognising Faces across Pose and Age }},
year = {2018},
pages = {67-74},
doi = {10.1109/FG.2018.00020},
url = {https://doi.ieeecomputersociety.org/10.1109/FG.2018.00020},
publisher = {IEEE Computer Society},
address = {Los Alamitos, CA, USA},
month =May}

@article{ElasticFace,
    title={ElasticFace: Elastic Margin Loss for Deep Face Recognition},
    author={Fadi Boutros and N. Damer and Florian Kirchbuchner and Arjan Kuijper},
    journal={2022 IEEE/CVF Conference on Computer Vision and Pattern Recognition Workshops (CVPRW)},
    year={2021},
    pages={1577-1586},
    doi={10.1109/CVPRW56347.2022.00164}}

@InProceedings{fenchel-young-losses,
  title = 	 {Learning Classifiers with Fenchel-Young Losses: Generalized Entropies, Margins, and Algorithms},
  author =       {Blondel, Mathieu and Martins, Andre and Niculae, Vlad},
  booktitle = 	 {Proceedings of the Twenty-Second International Conference on Artificial Intelligence and Statistics},
  pages = 	 {606--615},
  year = 	 {2019},
  editor = 	 {Chaudhuri, Kamalika and Sugiyama, Masashi},
  volume = 	 {89},
  series = 	 {Proceedings of Machine Learning Research},
  month = 	 {16--18 Apr},
  publisher =    {PMLR},
  url = 	 {https://proceedings.mlr.press/v89/blondel19a.html},
}

@inproceedings{f-divergence,
title={Beyond Reverse {KL}: Generalizing Direct Preference Optimization with Diverse Divergence Constraints},
author={Chaoqi Wang and Yibo Jiang and Chenghao Yang and Han Liu and Yuxin Chen},
booktitle={The Twelfth International Conference on Learning Representations},
year={2024},
url={https://openreview.net/forum?id=2cRzmWXK9N}
}

@inproceedings{L-Softmax,
author = {Liu, Weiyang and Wen, Yandong and Yu, Zhiding and Yang, Meng},
title = {Large-margin softmax loss for convolutional neural networks},
year = {2016},
publisher = {JMLR.org},
booktitle = {Proceedings of the 33rd International Conference on International Conference on Machine Learning - Volume 48},
pages = {507–516},
numpages = {10},
location = {New York, NY, USA},
series = {ICML'16}
}

@ARTICLE{KappaFace,
  author={Oinar, Chingis and M. Le, Binh and Woo, Simon S.},
  journal={IEEE Access}, 
  title={KappaFace: Adaptive Additive Angular Margin Loss for Deep Face Recognition}, 
  year={2023},
  volume={11},
  number={},
  pages={137138-137150},
  doi={10.1109/ACCESS.2023.3338648}}

@inproceedings{PartialFC-10-million,
  title={Partial fc: Training 10 million identities on a single machine},
  author={An, Xiang and Zhu, Xuhan and Gao, Yuan and Xiao, Yang and Zhao, Yongle and Feng, Ziyong and Wu, Lan and Qin, Bin and Zhang, Ming and Zhang, Debing and others},
  booktitle={Proceedings of the IEEE/CVF International Conference on Computer Vision},
  pages={1445--1449},
  year={2021}
}

@inproceedings{PartialFC-2-birds,
  title={Killing two birds with one stone: Efficient and robust training of face recognition cnns by partial fc},
  author={An, Xiang and Deng, Jiankang and Guo, Jia and Feng, Ziyong and Zhu, XuHan and Yang, Jing and Liu, Tongliang},
  booktitle={Proceedings of the IEEE/CVF conference on computer vision and pattern recognition},
  pages={4042--4051},
  year={2022}
}

@article{li2016renyi,
  title={R{\'e}nyi divergence variational inference},
  author={Li, Yingzhen and Turner, Richard E},
  journal={Advances in neural information processing systems},
  volume={29},
  year={2016}
}

@inproceedings{renyi1961measures,
  title={On measures of entropy and information},
  author={R{\'e}nyi, Alfr{\'e}d},
  booktitle={Proceedings of the fourth Berkeley symposium on mathematical statistics and probability, volume 1: contributions to the theory of statistics},
  volume={4},
  pages={547--562},
  year={1961},
  organization={University of California Press}
}

@inproceedings{f-divergence-novello,
author = {Novello, Nicola and Tonello, Andrea M.},
title = {f-divergence based classification: beyond the use of cross-entropy},
year = {2024},
publisher = {JMLR.org},
booktitle = {Proceedings of the 41st International Conference on Machine Learning},
articleno = {1560},
numpages = {26},
location = {Vienna, Austria},
series = {ICML'24}
}

@INPROCEEDINGS{ms1mv3,
  author={Deng, Jiankang and Guo, Jia and Zhang, Debing and Deng, Yafeng and Lu, Xiangju and Shi, Song},
  booktitle={2019 IEEE/CVF International Conference on Computer Vision Workshop (ICCVW)}, 
  title={Lightweight Face Recognition Challenge}, 
  year={2019},
  volume={},
  number={},
  pages={2638-2646},
  doi={10.1109/ICCVW.2019.00322}}

@ARTICLE{webface260m,
  author={Zhu, Zheng and Huang, Guan and Deng, Jiankang and Ye, Yun and Huang, Junjie and Chen, Xinze and Zhu, Jiagang and Yang, Tian and Du, Dalong and Lu, Jiwen and Zhou, Jie},
  journal={IEEE Transactions on Pattern Analysis and Machine Intelligence}, 
  title={WebFace260M: A Benchmark for Million-Scale Deep Face Recognition}, 
  year={2023},
  volume={45},
  number={2},
  pages={2627-2644},
  doi={10.1109/TPAMI.2022.3169734}}

@inproceedings{lfw,
  title={Labeled faces in the wild: A database forstudying face recognition in unconstrained environments},
  author={Huang, Gary B and Mattar, Marwan and Berg, Tamara and Learned-Miller, Eric},
  booktitle={Workshop on faces in'Real-Life'Images: detection, alignment, and recognition},
  year={2008}
}

@inproceedings{cfp-fp,
  title={Frontal to profile face verification in the wild},
  author={Sengupta, Soumyadip and Chen, Jun-Cheng and Castillo, Carlos and Patel, Vishal M and Chellappa, Rama and Jacobs, David W},
  booktitle={2016 IEEE winter conference on applications of computer vision (WACV)},
  pages={1--9},
  year={2016},
  organization={IEEE}
}

@article{cp-lfw,
  title={Cross-pose lfw: A database for studying cross-pose face recognition in unconstrained environments},
  author={Zheng, Tianyue and Deng, Weihong},
  journal={Beijing University of Posts and Telecommunications, Tech. Rep},
  volume={5},
  number={7},
  pages={5},
  year={2018}
}

@inproceedings{agedb,
  title={Agedb: the first manually collected, in-the-wild age database},
  author={Moschoglou, Stylianos and Papaioannou, Athanasios and Sagonas, Christos and Deng, Jiankang and Kotsia, Irene and Zafeiriou, Stefanos},
  booktitle={proceedings of the IEEE conference on computer vision and pattern recognition workshops},
  pages={51--59},
  year={2017}
}

@article{ca-lfw,
  title={Cross-age lfw: A database for studying cross-age face recognition in unconstrained environments},
  author={Zheng, Tianyue and Deng, Weihong and Hu, Jiani},
  journal={arXiv preprint arXiv:1708.08197},
  year={2017}
}

@inproceedings{ijb-b,
  title={Iarpa janus benchmark-b face dataset},
  author={Whitelam, Cameron and Taborsky, Emma and Blanton, Austin and Maze, Brianna and Adams, Jocelyn and Miller, Tim and Kalka, Nathan and Jain, Anil K and Duncan, James A and Allen, Kristen and others},
  booktitle={proceedings of the IEEE conference on computer vision and pattern recognition workshops},
  pages={90--98},
  year={2017}
}

@inproceedings{ijb-c,
  title={Iarpa janus benchmark-c: Face dataset and protocol},
  author={Maze, Brianna and Adams, Jocelyn and Duncan, James A and Kalka, Nathan and Miller, Tim and Otto, Charles and Jain, Anil K and Niggel, W Tyler and Anderson, Janet and Cheney, Jordan and others},
  booktitle={2018 international conference on biometrics (ICB)},
  pages={158--165},
  year={2018},
  organization={IEEE}
}

@inproceedings{resnet,
  title={Deep residual learning for image recognition},
  author={He, Kaiming and Zhang, Xiangyu and Ren, Shaoqing and Sun, Jian},
  booktitle={Proceedings of the IEEE conference on computer vision and pattern recognition},
  pages={770--778},
  year={2016}
}

@inproceedings{angularsparsemax,
  title={Angular sparsemax for face recognition},
  author={Chan, Chi-Ho and Kittler, Josef},
  booktitle={2020 25th International Conference on Pattern Recognition (ICPR)},
  pages={10473--10479},
  year={2021},
  organization={IEEE}
}

@inproceedings{mobilefacenets,
  title={Mobilefacenets: Efficient cnns for accurate real-time face verification on mobile devices},
  author={Chen, Sheng and Liu, Yang and Gao, Xiang and Han, Zhen},
  booktitle={Chinese conference on biometric recognition},
  pages={428--438},
  year={2018},
  organization={Springer}
}

@InProceedings{Voxceleb2,
  author       = "Chung, J.~S. and Nagrani, A. and Zisserman, A.",
  title        = {{VoxCeleb2: Deep} Speaker Recognition},
  booktitle    = "Proc. Interspeech",
  year         = "2018",
  pages     = {1086-1090},
}

@article{wang2024advancing,
  title={Advancing speaker embedding learning: {Wespeaker} toolkit for research and production},
  author={Wang, Shuai and Chen, Zhengyang and Han, Bing and Wang, Hongji and Liang, Chengdong and Zhang, Binbin and Xiang, Xu and Ding, Wen and Rohdin, Johan and Silnova, Anna and others},
  journal={Speech Communication},
  volume={162},
  pages={103104},
  year={2024},
  publisher={Elsevier}
}

@INPROCEEDINGS{xiang2019_margin-matters,
  author={Xiang, Xu and Wang, Shuai and Huang, Houjun and Qian, Yanmin and Yu, Kai},
  booktitle={2019 Asia-Pacific Signal and Information Processing Association Annual Summit and Conference (APSIPA ASC)}, 
  title={Margin Matters: Towards More Discriminative Deep Neural Network Embeddings for Speaker Recognition}, 
  year={2019},
  volume={},
  number={},
  pages={1652-1656},
  doi={10.1109/APSIPAASC47483.2019.9023039}}

@ARTICLE{huh2024_voxsrc-retrospective,
  author={Huh, Jaesung and Chung, Joon Son and Nagrani, Arsha and Brown, Andrew and Jung, Jee-weon and Garcia-Romero, Daniel and Zisserman, Andrew},
  journal={IEEE/ACM Transactions on Audio, Speech, and Language Processing}, 
  title={The {VoxCeleb} Speaker Recognition Challenge: A Retrospective}, 
  year={2024},
  volume={32},
  number={},
  pages={3850-3866},
  doi={10.1109/TASLP.2024.3444456}}

@INPROCEEDINGS{Thienpondt21:lm-finetuning,
  author={Thienpondt, Jenthe and Desplanques, Brecht and Demuynck, Kris},
  booktitle={ICASSP 2021 - 2021 IEEE International Conference on Acoustics, Speech and Signal Processing (ICASSP)}, 
  title={The {IdLab VoxSRC-20} Submission:{ Large} Margin Fine-Tuning and Quality-Aware Score Calibration in {DNN} Based Speaker Verification}, 
  year={2021},
  volume={},
  number={},
  pages={5814-5818},
  doi={10.1109/ICASSP39728.2021.9414600}}
}


\end{document}